# Associative Memory using Attribute-Specific Neuron Groups-1: Learning between Multiple Cue Balls

**Hiroshi Inazawa**

*Center for Education in Information Systems, Kobe Shoin Women's University[1],*
*1-2-1 Shinohara-Obanoyama, Nada Kobe 657-0015, Japan.*
**E-mail**: inazawa706@gmail.com

Abstract

In this paper, we present a new neural network model based on attribute-specific representations (e.g., color, shape, size), a classic example of associative memory. The proposed model is based on a previous study on memory and recall of multiple images using the Cue Ball and Recall Net (referred to as the CB-RN system, or simply CB-RN) [1]. The system consists of three components, which are C.CB-RN for processing color, S.CB-RN for processing shape, and V.CB-RN for processing size. When an attribute data pattern is presented to the CB-RN system, the corresponding attribute pattern of the cue neurons within the Cue Balls is associatively recalled in the Recall Net. Each image pattern presented to these CB-RN systems is represented using a two-dimensional code, specifically a QR code [2].

## 1. Introduction

In this paper, we present a new neural network model of associative memory that operates attribute-wise. This model is based on a previous study on the memory and recall of multiple images using the Cue Ball and Recall Net (referred to as the CB-RN system, or simply CB-RN) [1]. In this paper, we investigated whether an associative system capable of recalling patterns of other attributes when a single pattern of a given attribute is presented after learning can be realized using the CB-RN system. We have prepared three types of attributes: one for color, one for shape, and one for size. Additional attributes can be added as needed, and other kinds of attributes could also be used without issue. In the CB-RN system, we say that the components referrers to as C.CB-RN for color attributes, S.CB-RN for shape attributes, and V.CB-RN for size attributes. Since the model discussed here is a prototype, the representations of color, shape, and size are based on commonly used terms that exist in our everyday environment. It is entirely possible to extend this prototype model to practical-level objects; however, the number of elements required to construct each attribute would become enormous. For example, in Japan, there are 465 different colors recognized, and the same goes for shapes, so it is not realistic to apply this model to real objects from the start. For this reason, this paper introduce a theoretical model that works in principle. The fundamental operation of the model follows the well-known mechanism of associative memory [3–9]. For example, in terms of fruit attributes, if the attributes are "red," "green," "round," and "large," the association would lead to "watermelon." In other words, the model introduced here is designed to specify one attribute and then learn or recall other attributes that are associated with it. Each attribute name is represented as a pattern image. Representing attributes in this way dramatically expands the range of possible applications—for instance, even subtle qualities such as taste, smell, or other sensory experiences can be expressed within this framework. In this study, the pattern images are represented using QR codes [2], which are widely used as a form of two-dimensional code. In general, when humans recall a single memory, it often triggers a chain reaction in which related memories are successively brought to mind. This process of memory recall seems to be closely related to imagery. We believe that the model presented here may capture one aspect of this mechanism. In recent years, AI—particularly generative AI—has advanced significantly and reached a practical stage [10–18]. These developments are based on the field of deep learning, which builds on the techniques of multilayer neural networks studied during the late 1980s to early 1990s (the second wave of neural network research) [19–28]. On the other hand, memory-related models are generally considered to belong to the field of associative memory [3, 6–9, 27, 28]. However, models specifically focused on memory have not attracted as much attention as the learning models used in generative AI. In addition, it is worth noting that recent approaches have explored building memories using learning based on transformer architectures [29]. In Section 2, we describe the details of the model; Section 3 presents the results of the simulations, and Section 4 is devoted to conclusions and discussion.

---

[1] This was my affiliated institution until the end of March 2025, when I retired. Please note that as of April 2025, the university name has been changed from "Kobe Shoin Women's University" to "Kobe Shoin University."

## 2. Specification of the model

In this Section, we provide a detailed explanation of the model. The proposed model performs bidirectional memory learning between neurons in multiple clusters, called Cue Balls, (hereafter referred to as cue neurons[2]), neurons in a Recall Net consisting of many neurons representing images (hereafter referred to as recall neurons), and neurons within each Cue Ball. An overview of the CB-RN system is shown in Figure 1. In this system, when attributes such as color, shape, and volume are presented to the Recall Net, each attribute is represented as a QR code image as a simple character (e.g., "red"). In the actual simulation, the pixels of the QR code image are converted into digital values and then input into the neurons of the Recall Net. Also each QR code is made up of 116×116 pixels, resulting in 13,456 digital values.

---

[2] The "grandmother cell" hypothesis was once proposed, suggesting that a single neuron could be responsible for representing a specific memory.

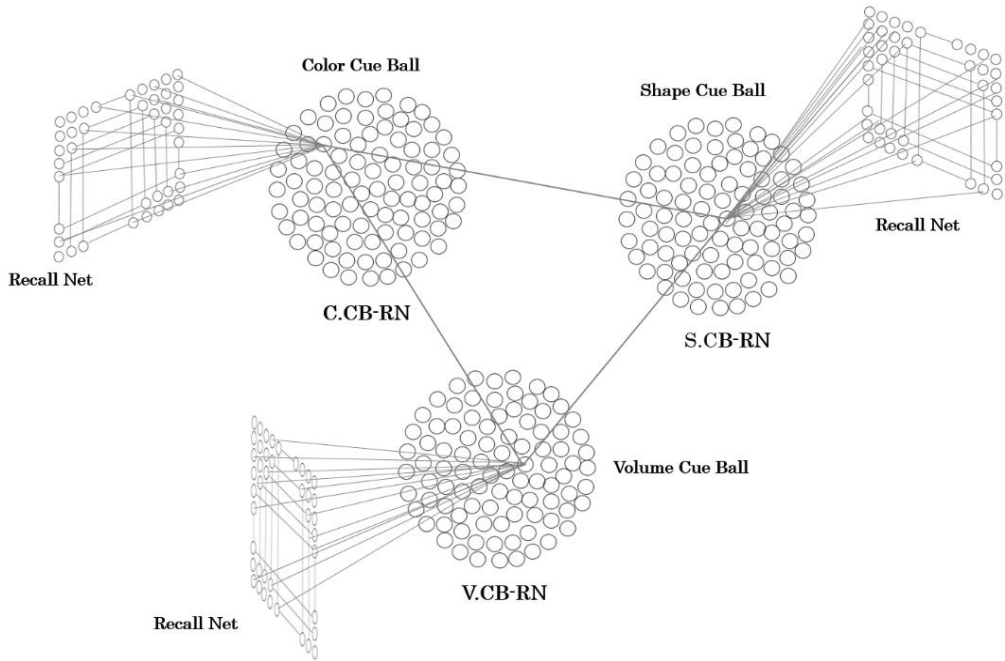

Figure 1. Schematic of the CB-RN system. Each Cue Ball is represented as a sphere, and one Recall Net is assigned to each Cue Ball. Note that a single Recall Net is sufficient for all Cue Balls; it is not necessary to assign one Recall Net to each Cue Ball. In this figure, a separate Recall Net is illustrated for each Cue Ball for the sake of clarity. All recall neurons within each Recall Net are connected to all cue neurons within the corresponding Cue Ball. There are no internal connections between recall neurons within the Recall Net. In addition, connections exist between cue neurons across different Cue Balls, whereas no connections are present among cue neurons within the same Cue Ball.

The model presented here is based on our previously published model; therefore, for detailed descriptions of its operational mechanisms, please refer to References [1]. Therefore, here we only introduce the basic operation formula. Each cue neuron within a Cue Ball is connected to all recall neurons in the Recall Net. In addition, that cue neuron is connected to all cue neurons in the other two Cue Balls. Note that all cue neurons have a threshold. In addition, no internal connections exist among cue neurons within the same Cue Ball and nor among recall neurons within the Recall Net. In principle, the number of cue neurons in the Cue Ball can be increased without limit, but in this paper, the number of cue neurons in the Cue Ball becomes to the number of prepared attribute data (number of image patterns). As mentioned above, the number of neurons (i.e., the dimension of the vector) of the Recall Net is unified to $116 \times 116$, forming a 13,456-dimensional vector. The operational process is divided into two phases: the learning phase and the recall phase. The learning phase is described first, followed by the recall phase. The general procedure of the learning operation is as follows: after presenting the pattern data to be learned (encoded) to the Recall Net, learning executes between one cue neuron in the corresponding Cue Ball—which is connected to all recall neurons within that Recall Net—and those recall neurons, through adjustment of the connection weights $w_{ji}^a$. Next, the outputs from all recall neurons in the Recall Net are fed back to the same cue neuron, and learning occurs in the cue neuron through the adjustment of its connection weights $v_{ij}^a$. This completes bidirectional memorization learning between the cue neuron and the recall neuron. This process is performed for three CB-RN systems corresponding to all attributes (color, shape, size). After training between the Recall Net and its corresponding Cue Ball is complete, learning occurs between the cues neurons for different Cue Balls. For example, learning occurs between the $k$th cue neuron in the Color Cue Ball and the $l$th cue neuron in the Style Cue Ball. Next, let me explain the general operation of the recall process. To check whether the recall neuron weights $w_{ji}^a$ have been successfully learned, we simply input the output of any cue neuron in the Cue Ball to all recall neurons in the Recall Net and examine the QR code pattern that appears in the Recall Net. In addition, the success or failure of learning $v_{ij}^a$" can be confirmed through the following process. After presenting the learned image pattern to the Recall Net, the output value of each recall neuron in that Recall Net is sent to all cue neurons in the corresponding Cue Ball. Each cue neuron that receives the input will produce an output, and if a cue neuron is found whose value is close to the learning value $\theta$ set during learning, it means that that cue neuron has learned the pattern in question. By normalizing the output of this cue neuron to 1.0 using a threshold function and feeding it back to the recall neurons of the Recall Net, the learned (presented) pattern is displayed in the Recall Net. Note that lowering this threshold allows multiple patterns that closely resemble the initially presented pattern to appear as candidates during recall. Finally, we discuss the association between the cue neurons of each Cue Ball. As described above, when a learned QR-code pattern is presented to a Recall Net, the corresponding cue neuron in the associated Cue Ball (for example, the $k$th neuron) produces an output. This output is sent to all cue neurons in another Cue Ball, where only previously trained cue neurons (e.g., the $l$th neuron) produce an output, indicating successful training.

Now, let's use mathematical formulas to describe the above learning and recall processes in more detail. The explanation will proceed through the learning process, followed by the recall process. In the learning process, the input-output relation from one cue neuron in any Cue Ball to all recall neurons in the Recall Net is expressed as follows:

$$y_j^a = w_{ji}^a x_i^a \quad , \tag{1}$$

where "$a$" is used to identify the attributes (color, style, volume, etc.) of each Cue Ball.

| $a, b \equiv C (= Color)$ , $S (= Style)$ , $V (= Volume)$ | | |
|---|---|---|
| $C (= Color)$ | $S (= Style)$ | $V (= Volume)$ |
| 0. red | 0. square | 0. extra − large |
| 1. orange | 1. circle | 1. large |
| 2. yellow | 2. oval | 2. medium |
| 3. green | 3. rectangle | 3. small − medium |
| 4. blue | 4. trapeziod | 4. small |
| 5. indigo | 5. triangle | 5. extra − small |
| 6. purple | 6. rhombus | 6. mini |

Table 1: Cue Ball Identification and Contents. The identification method for each Cue Ball is written at the top. Each attribute type is assigned a number ranging from 0 to 6. In practice, when used, these labels (e.g., the string "$red$") are converted into QR-code images.

When converting to a QR code, characters that explain each attribute (for example, "red") are converted into a QR-code. $x_i^a$ represents the output of the $i$th cue neuron, and yja represents the output of the $j$th recall neuron. $w_{ji}^a$ denotes the connection weight from the $i$th cue neuron to the $j$th recall neuron. Equation (1) is a commonly used mathematical expression; however, it is important to note that the summation over "$i$" is not performed. We uses the gradient descent method (GDM) [21-22] as the learning process. Based on this, the error function $E^a$ for the clue neuron is expressed as follows:

$$E^a \equiv \frac{1}{2} \sum_{j=0}^{M} \left( d_j^{ap} - y_j^a \right)^2 \quad , \tag{2}$$

where $d_j^{ap}$ denotes the element value of the QR code pattern presented to the Recall Net. $p$ is the pattern number, and $M$ is the number of recall neurons in one Recall Net ($M + 1 = 116 \times 116 = 13{,}456$). Furthermore, the update equation for the connection weight $w_{ji}^a$, as derived using the gradient descent method (GDM), is expressed as follows [1].

$$w_{ji}^a(t+1) = w_{ji}^a(t) + \Delta w_{ji}^a(t) \tag{3}$$

$$\Delta w_{ji}^a(t) = -\varepsilon_W \frac{\partial E^a}{\partial w_{ji}^a} = \varepsilon_W \left( d_j^{ap} - y_j^a(t) \right) x_i^a(t) \quad , \tag{4}$$

where, $t$ denotes the number of updates, and $\varepsilon_W$ is the learning rate, which is common to all recall neurons.

In addition, $\varepsilon_W$ is set to 1. By applying the input–output relation defined in Equation (1) along with the learning rules for the connection weights described in Equations (3) and (4), it can be confirmed that $w_{ji}^a$ is capable of reliably learning the presented patterns. Below, we show this analytically. First, compute Equation (1) at "$t + 1$" using Equations (3) and (4).

$$y_j^a(t+1) = w_{ji}^a(t+1)x_i^a(t+1)$$
$$= \left(w_{ji}^a(t) + \left(d_j^{ap} - w_{ji}^a x_i^a(t)\right)\right)x_i^a(t+1). \tag{5}$$

Furthermore, by setting "$x_i^a(t) = x_i^a(t+1) \equiv 1$", Equation (5) can be rewritten as follows:

$$y_j^a(t+1) = d_j^{ap}. \tag{6}$$

Therefore, after training, "$y_j^a(t+1)$" will exactly match the element values of the presented letter pattern.

Next, let's examine the learning process of the cue neuron connection weights $v_{ij}^a$. The input–output relation from the recall neurons to the cue neurons is described as follows:

$$x_i^a = f(q_i^a) = f\left(\sum_{j=0}^{M} v_{ij}^a y_j^a\right) = \begin{cases} 0 & \text{for } q_i^a < D \\ 1 & \text{for } q_i^a \geq D \end{cases}, \tag{7}$$

where $q_i^a$ denotes the output of the $i$th cue neuron before thresholding, and $y_j^a$ represents the input from the $j$th recall neuron. In this case, the value of $y_j^a$ is obtained from equation (1) with "$x_i^a = 1$" after learning $w_{ji}^a$. $v_{ij}^a$ is the connection weight from the $j$-th recall neuron to the $i$-th cue neuron, and f is the threshold function. The threshold $D$ is the same for all cue neurons of each Cue Ball. The learning performs using the same GDM as in the $w_{ji}^a$ learning. In this case, the error function for the cue neuron is as follows:

$$e^a \equiv \frac{1}{2}\sum_{i=0}^{L}(\theta - q_i^a)^2, \tag{8}$$

where $\theta$ is an arbitrary value (constant) set as the learning value of $v_{ij}^g$, and we sets the same for all cue neurons in each Cue Ball. Furthermore, L is equal to the number of attributes within each Cue Ball minus one; thus, "L=6" for each case. Using these values, the weight update equation derived from the GDM is given as follows.

$$v_{ij}^a(t'+1) = v_{ij}^a(t') + \Delta v_{ij}^a(t') \ , \tag{9}$$

$$\Delta v_{ij}^a(t') = -\varepsilon_V \frac{\partial e}{\partial v_{ij}^a} = \varepsilon_V \big(\theta - q_i^a(t')\big) y_j^a(t') \ , \tag{10}$$

where $t'$ denotes the number of update, and $\varepsilon_V$ represents the learning rate, which is common to all cue neurons. As with $\varepsilon_W$, $\varepsilon_V$ is set to 1. The input value $y_j^a$ is obtained by equation (1) after learning $w_{ji}^a$. Similar to the learning process for $w_{ji}^a$, $v_{ij}^a$ can be reliably learned by applying the input–output relation given in Equation (7) together with the weight update rules in Equations (9) and (10). In this case, $y_j^a$ in equation (1) is calculated as "$x_i^a = 1.0$". Substituting $v_{ij}^a$ in equations (9) and (10) into the equation for the intermediate value $q_i^a$ in equation (7), the following relation is obtained.

$$\begin{aligned} q_i^a(t'+1) &= \sum_{j=0}^M v_{ij}^a(t'+1) y_j \\ &= \sum_{j=0}^M v_{ij}^a(t') y_j^a \left(1 - \sum_{k=0}^M (y_k^a)^2\right) + \theta \sum_{j=0}^M (y_j^a)^2 \ , \end{aligned} \tag{11}$$

where we consider that $y_j^a$ in equation (1) is equal to $d_j^{ap}$, and set the following normalization condition for $d_j^{ap}$.

$$\sum_{j=0}^M (y_j^a)^2 = \sum_{j=0}^M (d_j^{ap})^2 \equiv 1 \tag{12}$$

Therefore, using equation (12), equation (11) becomes as follows:

$$q_i^a(t'+1) = \theta \ . \tag{13}$$

In this way, $q_i^a(t'+1)$ coincides with the learned value $\theta$ after learning $v_{ij}^a$.

Finally, we explain the learning between the cue neurons of each Cue Ball. During this learning, it is assumed that each Cue Ball and Recall Net neuron has already been learned. For example, when a pattern is presented to the Recall Net, the output $x_k^a$ of the $k$th cue neuron in Cue Ball "$a$" is used to learn the $l$th cue neuron in Cue Ball "$b$." In this case, the input/output relationship between the cue neurons has the same format as equation (7) and is as follows:

$$x_l^b (\equiv z_l^b) = f(q_l^b) = f(u_{lk}^b z_k^a) = \begin{cases} 0 & for\ q_l^b < D \\ 1 & for\ q_l^b \geq D \end{cases}, \quad (14)$$

In this Section, the notation "$x_k^a, x_l^b$" is used instead of "$z_k^a, z_l^b$" to represent the output values resulting from the processing between the Recall Net and the Cue Ball. Note that the data content is the same. $z_k^a$ is the input value "$z_k^a (\equiv x_k^a)$" using the output value "$x_k^a = 1.0$" of the kth cue neuron of Cue Ball "$a$" based on the pattern presented to the Recall Net. $u_{lk}^b$ is the connection weight from the kth cue neuron of Cue ball "$a$" to the lth cue neuron of Cue ball "$b$". $f$ is a threshold function, and $D$ is a common threshold value for the cue neurons of each Cue Ball. In addition, by interchanging "$a \rightarrow b$" and "$b \rightarrow a$" in Equation (14), the expression represents the relation between the output values of Cue Ball "$a$" and the input values of Cue Ball "$b$" (see Table 1 for $a$ and $b$). The relation between the variables are becoming more complex, so a schematic diagram is shown in Figure 2.

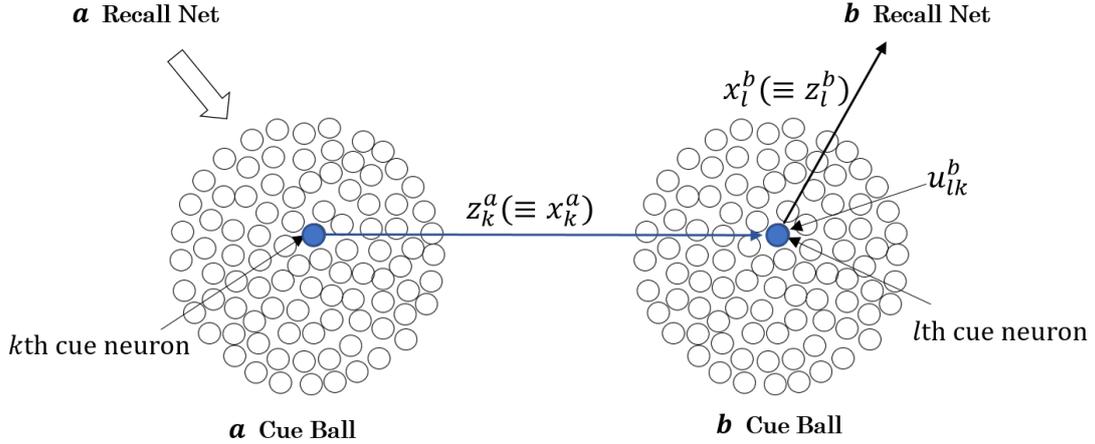

Figure 2: Learning of cue neurons between Cue Balls. The input and output variables and coupling weights of the cue neurons in the Cue Balls in $a$ and $b$ (=Color, Style, Volume) are shown. When a pattern is presented to Recall Net, the $k$th cue neuron of Cue Ball "$a$" produces an output "$z_k^a (\equiv x_k^a)$," which in turn activates the learned cue neurons within Cue Ball "$b$," resulting in their output $x_l^b$. This causes the learned QR-code pattern to be displayed in the Recall Net.

Next, we show the learning formula for the cue neuron of Cue Ball "$b$." First, the error function for the cue neuron of Cue Ball "$b$" is as follows:

$$\eta^b \equiv \frac{1}{2}\sum_{l=0}^{L}(\theta - q_l^b)^2 \quad , \tag{15}$$

where $\theta$ is an arbitrary value (constant value) set as the learning value of $u_{lk}^b$. $\theta$ is the same as in equation (8), and the same is true for the coupling constants between each cue neuron. Using these, the update formula for the coupling weights obtained by GDM is as follows:

$$u_{lk}^b(t+1) = u_{lk}^b(t) + \Delta u_{lk}^b(t) \quad , \tag{16}$$

$$\Delta u_{lk}^b(t) = -\lambda_{CB}\frac{\partial \eta^b}{\partial u_{lk}^b} = \lambda_{CB}\left(\theta - q_l^b(t)\right)z_k^a(t) \quad , \tag{17}$$

where $t$ is the number of updates, and $\lambda_{CB}$ is the learning rate. $\lambda_{CB}$ is set to 1, just like $\varepsilon_V$ and $\varepsilon_W$. By using the input–output relationship defined in Equation (14) together with the learning rules for the coupling weights given in Equations (16) and (17), the parameter $u_{lk}^b$ can be reliably learned. In this case, since the Recall Net is presented with a trained pattern image, the output value $x_k^a$ of the $k$-th neuron is 1.0, i.e. $z_k^a \equiv x_k^a = 1.0$. Note that the output values of all cue neurons other than the $k$th neuron in this Cue Ball are 0.0, where by using equations (14), (16), and (17), the following relationship is obtained:

$$\begin{aligned}q_l^b(t+1) &= u_{lk}^b(t+1)z_k^a \\ &= u_{lk}^b(t) + \theta - q_l^b(t) = \theta\end{aligned} \quad . \tag{18}$$

In this way, $q_l^b(t+1)$ coincides with the learned value $\theta$ after learning $u_{lk}^b$.

3. Simulation and Results

In this Section, we verify the operation of the learning and recall processes described in Section 2 through simulation. The following is a repeat of Section 2, but written down again before running the simulation. As explained in Chapter 2, the pattern data used in the simulation are expressed as characters according to its attributes, which are converted into a QR code (see the Appendix for pattern images). Each QR-code image is composed of $116 \times 116 = 13,456$ pixels, which are used as a 13,456 dimensional digital value vector. The specific system configuration consists of three components: the "C.CB-RN," which represents the attribute of color through the Color Cue Ball and the Recall Net; the "S.CB-RN," which represents the attribute of shape through the Style Cue Ball and the Recall Net; and the "V.CB-RN," which represents the attribute of volume through the Volume Cue Ball and the Recall Net. During the training process, one image is extracted from each attribute group (see Table 1) and mutual training is performed between the Recall Net "$w_{ji}^a$" and the Cue Ball "$v_{ij}^a$". After that, training is performed on the Cue neurons between each Cue Ball. In the learning process of cue neurons between Cue Balls, a previously learned image pattern is first presented to the Recall Net to one of the Cue Balls. In this state, the cue neurons corresponding to the learned pattern in that Cue Ball are generating outputs. On this state, the learning among the cue neurons is performed in accordance with Equations (14)–(17). The same procedure is then applied to the other Cue Ball and the Recall Net.

Below, the simulation algorithms for the learning and recall processes are outlined. Let's begin with the learning process of $w_{ji}^b$.

[Algorithm 1 - Learning $w_{ji}^a$]
1. Specify each Cue Ball
2. Setting each input/output variable, connection weight initialization, and learning rate：

   $x_i^a = 1.0,\ y_j^a = 1.0,\ w_{ji}^a = 0.0, \varepsilon_W = 1.0$

3. For each pattern presented to the Recall Net, apply procedures (3) and (4) above to Equations (4) and (5), and use the resulting updates to be learned the synaptic (connection) weights.
4. Save the value of $w_{ji}^a$ after learning to a file
5. Repeat steps 2 to 4 above while changing attribute groups.

The learning algorithm for $v_{ij}^a$ is as follows.

[Algorithm 2 - Learning $v_{ij}^a$]
1. Specify each Cue Ball
2. Initialization of each input/output variable and connection weight, setting of learning coefficient, learning value, and threshold:
   $x_i^a = 1.0,\ q_i^a = 0.0,\ y_j^a = 1.0,\ v_{ij}^a = 0.0,\ \varepsilon_V = 1.0,\ \theta = 100.0,\ D = 72.0$
   ※The threshold value "$D = 72.0$" has been determined based on the results of the simulation.
3. Calculate $y_j^a$ using the output of cue neurons for each Cue Ball, $x_i^a$, and equation (1):
   $w_{ji}^a$ has been learned
4. Calculate $q_i^a$ in equation (8) for cue neurons of each Cue Ball: Use $y_j^a$ from step 3
5. The connection weights in equation (10) are learned using $q_i^a$ and $y_j^a$ calculated in step "3".
6. Save the value of $v_{ij}^a$ after learning to a file
7. Repeat steps 2 to 5 above while changing attribute groups.

Finally, we explain the algorithm for the learning process of the cue neurons between each Cue Ball.

[Algorithm 3 - Learning $u_{kl}^b$]

1. Specify one of the attribute groups " cmb = 0,1,2" to present (each number specifies two Cue Balls):
$$\text{cmb} = 0 \rightarrow a = \text{Color}, b = \text{Style},$$
$$\text{cmb} = 1 \rightarrow a = \text{Style}, b = \text{Volume},$$
$$\text{cmb} = 2 \rightarrow a = \text{Volume}, b = \text{Color}$$

2. Initialization of each input/output variable and connection weight, setting of learning coefficient, learning value, and threshold:
$x_i^a = 1.0, \ q_i^a = 0.0, \ y_j^a = 1.0, \ v_{ij}^a = 0.0, \ \varepsilon_V = 1.0, \ \theta = 100.0, \ D = 72.0$
※The threshold value "$D = 72.0$" has been determined based on the results of the simulation.

3. Select one cue neuron each in Cue Ball "$a$" and Cue Ball "$b$": Let these be the $k$th neuron and the lth neuron

4. After presenting one of the learned patterns to the Recall Net associated with Cue Ball "$a$," the output of the $k$th neuron, $z_k^a (\equiv x_k^a)$, is used to calculate the output of the lth neuron, $x_k^a (\equiv z_l^b)$, for Cue Ball "$b$," and then $q_l^b$.

5. The connection weights of equations (17) and (18) are learned using $q_l^b$ and $z_k^a$ calculated in step 4.

6. Save the value of $u_{kl}^b$ after learning to a file

7. Swap Cue Ball "a" and Cue Ball "b" and perform steps "2" to "6".

8. Return to step "1" and repeat the above steps for all Cue Ball combinations

Next, we describe the recall process that follows after each learning process. This explanation focuses specifically on a cue neuron (the $k$th neuron) within Cue Ball "$a$" in each attribute group. When the Recall Net is presented with the QR-code image learned by the kth neuron, it produces an output of $x_k^a (\equiv z_k^a = 1.0)$. Using this output, the responses of all cue neurons in the learned Cue Ball "$b$" are examined. Here, the neuron exhibiting the maximum output value, $xx_l^b (\equiv z_l^b = 1.0)$, is substituted into Equation (1) to calculate $y_l^b$, where this display the QR-code pattern to the Recall Net. The above-described algorithm is described below.

[Algorithm 4 - Recall by learning between cue neurons]

1. Specify one of the attribute groups " cmb = 0,1,2" to be presented (each number specifies

two Cue Balls): See the learning algorithm for $w_{ji}^b$

2. Present the QR-code pattern learned by an arbitrary cue neuron in Cue Ball "$a$" to the Recall Net.
3. In this process, the output value $x_k^a$ ($= z_k^a$) of the cue neuron for Cue Ball "$a$" is input to the cue neuron for Cue Ball "$b$," and the resulting output $z_l^b$ is calculated according to equation (15).
4. This output value "$z_l^b$ ($= x_l^b$)" is used in equation (1) to calculate $y_j^b$.
5. Check the QR-code pattern displayed in Recall Net using this output value $y_j^b$

In the following, we examine whether learning and recall operate as intended under these algorithms. After learning $w_{ji}^a$ and $v_{ij}^a$ according to Algorithms 1 and 2, we present specific training patterns for each attribute group (color, style, and volume) to the Recall Net. Figure 2a shows the output values $q_i^a$ of all cue neurons (0th to 6th) at this time, where the color Cue Ball is presented with pattern 1 (associated with neuron 0), the style Cue Ball with pattern 4 (associated with neuron 3), and the volume Cue Ball with pattern 7 (associated with neuron 6). The output value $q_i^a$ at this time is shown in Figure 3.

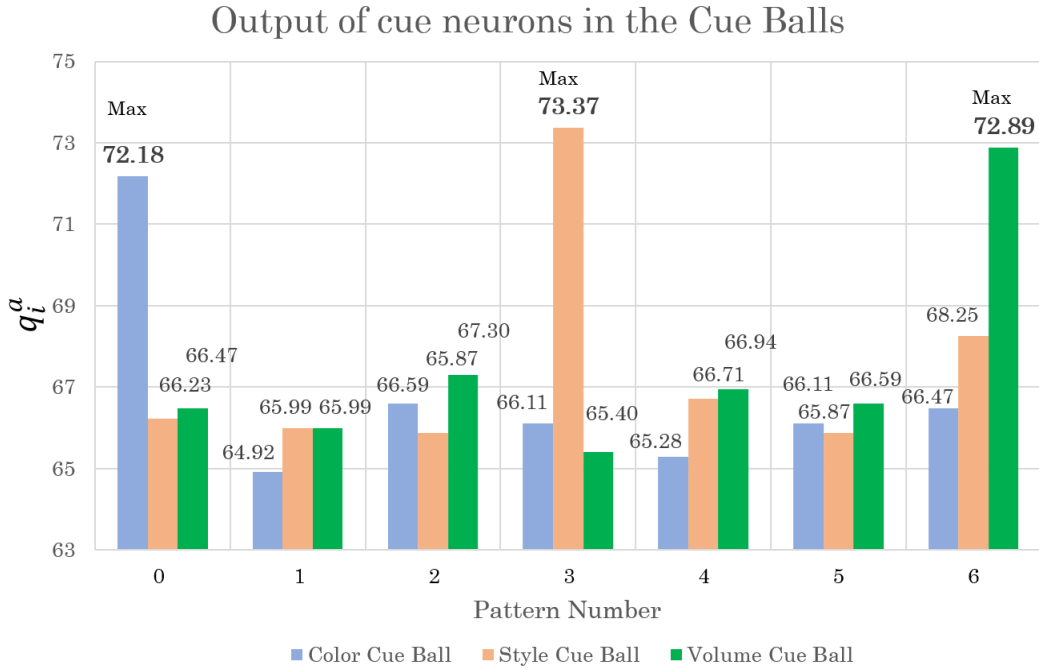

Figure 3: Output patterns of cue neurons in each Cue Ball. In the Color Cue Ball, the 0th neuron, in the Style Cue Ball, the 3rd neuron, and in the Volume Cue Ball, the 6th neuron are presented to the Recall Net. In the Color Cue Ball, the 0th neuron produces the largest output value $q_i^a$, resulting in "$x_{i=0}^{Color} = 1.0$," which confirms that learning has been successfully completed. In the Style Cue Ball, the 3rd neuron, and in the Volume Cue Ball, the 6th neuron produce the highest output values $q_i^a$,

resulting in "$x_{i=3}^{Style} = 1.0$" and "$x_{i=6}^{Volume} = 1.0$," respectively. These results also confirm that learning has been successfully completed.

For the color Cue Ball, 0th neuron outputs the highest output value ($q_i^{Color} = 72.65$). Applying the threshold function results in "$x_{i=0}^{Color} = 1.0$," indicating that pattern 1 has been correctly recalled. For the color Style Ball, 3rd neuron outputs the highest output value ($q_i^{Style} = 73.37$). Applying the threshold function results in "$x_{i=3}^{Style} = 1.0$," indicating that pattern 4 has been correctly recalled. For the color Volume Ball, 6th neuron outputs the highest output value ($q_i^{Volume} = 72.89$). Applying the threshold function results in "$x_{i=6}^{Volume} = 1.0$," indicating that pattern 4 has been correctly recalled. These indicate that learning is proceeding as expected. From these simulation results, since the maximum $q_i^a$ values of the three leaned cue neurons exceed 72.0, the threshold has been set to D=72.0.

Finally, let's run Algorithm 4. Attribute groups are executed in the order 0, 1, 2. The cue neuron numbers specified within each group are as follows: in Group 0, the 0th and 3rd neurons; in Group 1, the 3rd and 6th neurons; and in Group 2, the 6th and 1st neurons. For example, if the attribute group is 0 (color Cue Ball and style Cue Ball), then for color Cue Ball, the 0th neuron is selected, and for style Cue Ball, the 3rd neuron is selected.

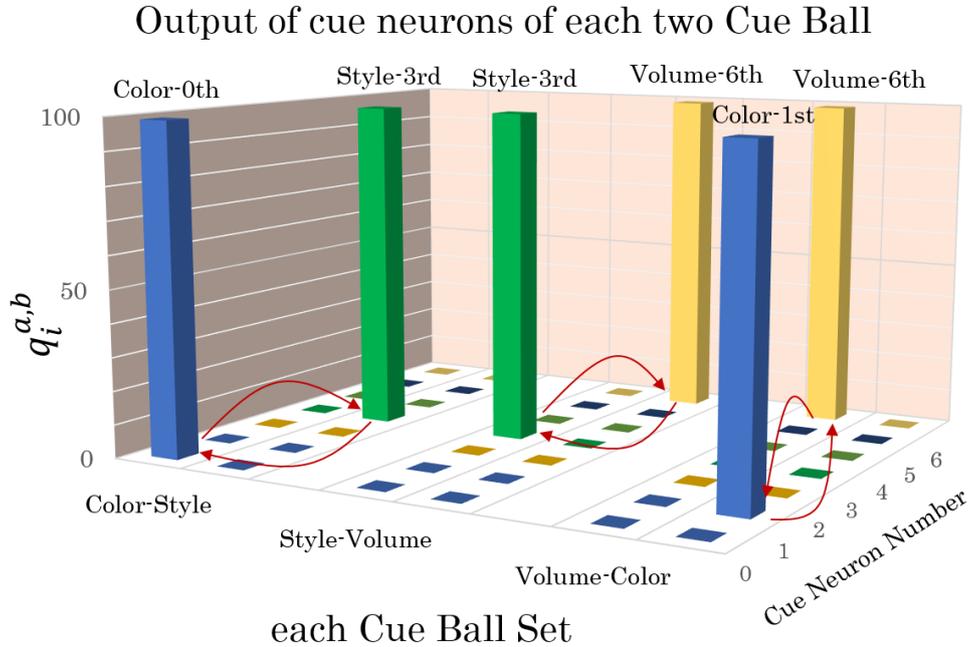

Figure 4: Output of learned cue neurons between each Cue Ball. The outputs of the cue neurons in each Cue Ball are 0 and 99. In Color-Style, when the 0th neuron of Color Cue Ball outputs "$z_{i=0}^{Color} = 1.0$", only the 3rd neuron of Style Cue Ball outputs. Conversely, if the 3rd neuron of Style Cue Ball outputs an output, only the 0th neuron of Color Cue Ball will have an output. Similarly, in Style-Volume, it is the third neuron in Style Cue Ball and the sixth neuron in Volume Cue Ball,

and in Volume-Color, it is the sixth neuron in Volume Cue Ball and the first neuron in Color Cue Ball.

Figure 4 shows the output of the cue neuron for one Cue Ball and the output of the cue neuron for the other Cue Ball after the cue neuron between each Cue Ball has completed learning. Learning was conducted reciprocally between specific pairs of neurons for each Cue Ball combination: between the 0th neuron of the Color Cue Ball and the 3rd neuron of the Style Cue Ball for the Color–Style combination; between the 3rd neuron of the Style Cue Ball and the 6th neuron of the Volume Cue Ball for the Style–Volume combination; and between the 6th neuron of the Volume Cue Ball and the 2nd neuron of the Color Cue Ball for the Volume–Color combination. As a result, in the Color–Style combination, when the 0th neuron of the Color Cue Ball produces an output "$z_{i=1}^{Color} = 1.0$," only the learned 3rd neuron of the Style Cue Ball becomes active. In the Style–Volume combination, when the 3rd neuron of the Style Cue Ball produces an output, only the learned 6th neuron of the Volume Cue Ball becomes active. In the Volume–Color combination, when the 6th neuron of the Volume Cue Ball produces an output, only the learned 2nd neuron of the Color Cue Ball becomes active. The simulation results showed that all neurons with large outputs were the corresponding learned neurons, each of which produced an output value of 99. Furthermore, the outputs of all other untrained neurons were zero. These results can confirm that both learning and recall worked as intended.

## 4. Conclusions and Discussions

This paper has introduced a new model of associative memory that operates for each attribute belonging to multiple attribute-specific categories. This model is based on a series of previously published studies [1] on image memory and recall using the CB-RN system. More specifically, this model allows for associative recall of the pattern of other attributes when a pattern of a particular attribute is presented after learning. Three types of attributes were prepared—color, shape, and size—each treated as an image pattern. For the model used here, we prepared a small set of prototypes representing color, shape, and size, using common everyday terms. The mode of operation follows the well-established patterns of associative memory [3–9]. To represent attributes as images, we used QR codes [2], which are widely employed as two-dimensional codes. As in this study, representing attributes as characters and then representing them as image patterns dramatically expands the range of potential applications. For example, even subtle tastes, odors, or other abstract sensory experiences can be represented. In Section 3, we conducted simulations on the learning and recall processes of the CB-RN system described in Section 2. As shown in Figures 3 and 4, the results indicate that learning and recall performed as expected. By normalizing the element values $d_j^{ap}$ of the image patterns (so that the sum of squares equals 1.0), it was confirmed that the learning of $v_{ij}^a$ is reliably executed, as shown in Equation

(13) of section 2. Furthermore, as shown in Equation (18), it was confirmed that $u_{lk}^b$ can also be learned reliably. Note that throughout the simulation, a threshold was set to normalize the intermediate output $q_i^a$ of the cue neuron to 1.0, consistent with this model.

In human memory recall, it is common that remembering a single item triggers the sequential recall of related memories, much like a chain reaction. In particular, memory recall might appear to be closely related to images. We suggest that the model presented here may represent one aspect of such an underlying mechanism.

It is generally estimated that humans can recall approximately five associative memories at a time 30-32. As a next step, we plan to extend the current model to construct a system capable of sequentially recalling associations across five different images. Additionally, in this study, we standardized all images to a resolution of $116 \times 116$ pixels, but in the future we hope to extend the model to handle images of various sizes.

# References

[1] Inazawa, H, "Hetero-Correlation-Associative Memory with Trigger Neurons: Accumulation of Memory through Additional Learning in Neural Networks", Complex Systems, 27, Issue 2, 187-197, 2018.; Inazawa, H, "An associative memory model with very high memory rate: Image storage by sequential addition learning". arXiv:2210.03893, https://doi.org/10.48550/arXiv.2210.03893,2022.; Inazawa, H,. "The Method for Storing Patterns in Neural Networks-Memorization and Recall of QR code Patterns-", 2025. arXiv:2504.06631, https://doi.org/10.48550/arXiv.2504.06631.

[2] DENSO WAVE. "QR Code". Available at: https://www.denso-wave.com/qrcode/

[3] Kohonen, T. Correlation Matrix Memories. *IEEE Trans., C-21,* 1972 353-359.; Kohonen, T. *Self-Organization and Assocoative Memory*, Springer, 1984.

[4] Anderson, J.R and Bower, G.H., *Human Associative Memory*, Psychology Press, 1974.

[5] Nakano, K., *Learning Process in a Model of Associative Memory*, Springer, 1971.

[6] Amari, S. & Maginu, K. Statistical Neurodynamics of Associative Memory. *Neural Networks, Vol. 1,* 1988 63-73.

[7] Hopfield, J.J. Neural Networks and Physical Systems with Emergent Collective Computational Abilities. *Proc. of the National Academy of Science USA 79*: 1982 2254-2258. ;Hopfield JJ, Tank DW (1985) Neural computation of decisions in optimization problems. Biol Cybern 52:141–152, 1985.

[8] Yoshizawa, S., Morita, M. & Amari, S. Capacity of Associative Memory Using a Nonmonotonic Neuron Model. *Neural Networks, Vol. 6,* 1993 167-176.

[9] Amit, D.J., Gutfreund, H. & Sompolinsky, H. Storing Infinite Numbers of Patterns in a Spin-Glass Model of Neural Networks. *Phys. Rev. Lett., 55,* 1985 1530-1533.

[10] Hinton, G.E. & Salakhutdinov, R. Reducing the Dimensionality of Data with Neural Networks. *Science, Vol. 313*, 2006 504-507.

[11] Hinton, G.E., Osindero, S. & Teh, Y. A Fast Learning Algorithm for Deep Belief Nets. *Neural Computation, 18*: 2006 1527-1544.

[12] Bengio, Y., Lamblin, P., Popovici, D. & Larochelle, H. Greedy Layer-Wise Training of Deep Networks. *In Proc. NIPS,* 2006.

[13] Lee, H., Grosse, R., Ranganath, R. & Ng, A.Y. Convolutional Deep Breif Networks for Scalable Unsupervised Learning of Hierarchical Representations. *In Proc. ICML,* 2009.

[14] Krizhevsky, A., Sutskever, I. & Hinton, G.E. ImageNet Classification with Deep Convolutional Neural Networks. *In Proc. NIPS,* 2012.

[15] Le, Q.V., Ranzato, M., Monga, R., Devin, M., Chen, K., Corrado, G.S., Dean, J. & Ng, A.Y. Building High-Level Features Using Large Scale Unsupervised Learning. *In Proc. ICML,* 2012.

[16] Krizhevsky, A., Sutskever, I. & Hinton, G.E. ImageNet Classification with Deep Convolutional Neural Networks. *In Proc. NIPS,* 2012.

[17] A. Vaswani, N. Shazeer, N. Parmar, J. Uszkoreit, L. Jones, A.N. Gomez, L. Kaiser and I. Polosukhin, "Attention Is All You Need," 2017.

[18] J.W. Rae, A. Potapenko, S.M. Jayakumar and T. Lillicrap, "*Compressive Transformers for Long-Range Sequence Modeling,*" 2020.

[19] Rumelhart, D.E., McCleland, J.L. & the PDP Research Group. Parallel Distributed Processing: *Explorations in the Microstructure of Cognition, Vol. 1: Foundations. Cambridge: MIT Press,* 1986.

[20] Fukushima, K. & Miyake, S. Neocognitron: A new Algorithm for Pattern Recognition Tolerant of Deformations and Shifts in Position. *Pattern Recognition, 15:* 1982 455-469.

[21] Widrow, B. & Hoff, M.E., Adaptive Switching Circuits. *In 1960 IRE WESCON Convention Record, Part 4,* 96-104. *New York: IRE. Reprinted in Anderson and Rosenfeld [1988],* 1960.

[22] Rescorla, R.A. & Wagner, A.R., A Theory of Pavlovian Conditioning: The Effectiveness of Reinforcement and Nonreinforcement. *In classical Conditioning II: Current Research and Theory, eds. Bloch, A.H. & Prokasy, W.F., 64-69, New York: Appleton-Century-Crofts,* 1972.

[23] LeCun, Y., Boser, B., Denker, J.S., Henderson, D., Howard, R.E., Hubard, W. & Jackel, L.D. Backpropagation Applied to Handwritten Zip Code Recognition. Neural Computation, 1(4): 1989 541-551.

[24] Tsutsumi, K. Cross-Coupled Hopfield Nets via Generalized-Delta-Rule-Based internetworks. *In Proc. IJCNN90-San Diego II,* 1990 *259-265.*

[25] Simard, P.Y., Steinkraus, D. & Platt, J. Best Practice for Convolutional Neural Networks Applied to Visual Document Analysis. *In Proc. ICDAR,* 2003.

[26] LeCun., Y, Bottou, L., Bengio, Y. & Haffner, P. Gradient-Based Learning Applied to Document Recognition. *Proceedings of the IEEE, 86(11):* 1998 2278-2324.

[27] Srebro, N. & Shraibman, A. Rank, Trace-Norm and Max-Norm. *In Proc. 18th Annual Conference on Learning Theory, COLT 2005,* 2005 545-560, Springer.

[28] Srivastava, N., Hinton, G.E., Krizhevsky, A., Sutskever, I. & Salakhutdinov, R. Dropout: A Simple Way to Prevent Neural Networks from Overfitting. Journal of Machine Learning Research, 15: 2014 1929-1958.

[29] Marcella Cornia, Matteo Stefanini, Lorenzo Baraldi, Rita Cucchiara, "Meshed-Memory Transformer for Image Captioning", 2020, arXiv:1912.08226, https://doi.org/10.48550/arXiv.1912.08226

[30] Nelson Cowan (2001) "The magical number 4 in short-term memory: A reconsideration of mental storage capacity"*,* Behavioral and Brain Sciences, 24(1), 87–114. Cambridge University Press &

Assessment+2ResearchGate+2; DOI: 10.1017/S0140525X01003922 Cambridge University Press & Assessment

[31] Nelson Cowan "The Magical Mystery Four: How Is Working Memory Capacity Limited, and Why", Current Directions in Psychological Science, 19(1), 2010, 51–57.

[32] Cowan, N., et al. "Theory and measurement of working memory capacity limits", The psychology of learning and motivation, 2008.

## Appendix

All QR code patterns used in this paper are listed below. The file name at the bottom of each pattern, such as "red" etc., is a string of characters written inside as a QR code.

| Attributes of Color Cue Ball | | | | | | |
|---|---|---|---|---|---|---|
| 0th | 1st | 2nd | 3rd | 4th | 5th | 6th |
| 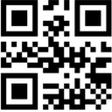 | 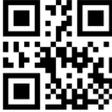 | 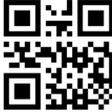 | 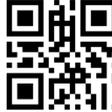 | 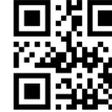 | 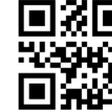 | 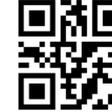 |
| red | orange | yellow | green | blue | indigo | purple |
| **Attributes of Style Cue Ball** | | | | | | |
| 0th | 1st | 2nd | 3rd | 4th | 5th | 6th |
| 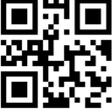 | 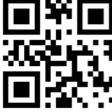 | 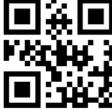 | 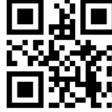 | 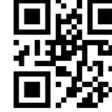 | 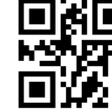 | 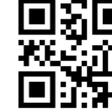 |
| square | circle | oval | rectangle | trapezoid | triangle | rhombus |
| **Attributes of Volume Cue Ball** | | | | | | |
| 0th | 1st | 2nd | 3rd | 4th | 5th | 6th |
| 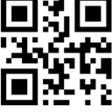 | 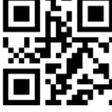 | 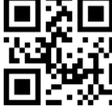 | 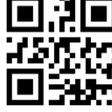 | 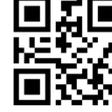 | 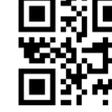 | 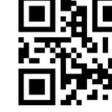 |
| extra-large | large | medium | small-medium | small | extra-small | mini |